%% file: PaperCameraReady.tex
\documentclass[10pt, twocolumn, letterpaper]{article}

\usepackage[pagenumbers]{cvpr} 

\usepackage{graphicx}
\usepackage{amsmath}
\usepackage{amssymb}
\usepackage{booktabs}

\usepackage{amsfonts}        
\usepackage{multicol} 
\usepackage{multirow} 
\usepackage{makecell} 
\usepackage{placeins} 
\usepackage[linewidth=1pt]{mdframed}
\usepackage{algpseudocode}
\usepackage{algorithm}

\usepackage{textcomp} 

\usepackage{color}
\usepackage{xcolor}

\definecolor{jiapeng}{rgb}{1.0,0.0,0.0}

\usepackage[pagebackref,breaklinks,colorlinks]{hyperref}

\usepackage[capitalize]{cleveref}
\crefname{section}{Sec.}{Secs.}
\Crefname{section}{Section}{Sections}
\Crefname{table}{Table}{Tables}
\crefname{table}{Tab.}{Tabs.}

\def\cvprPaperID{3718} 
\def\confName{CVPR}
\def\confYear{2023}

\begin{document}

\title{RGBD2: Generative Scene Synthesis via Incremental \\ View Inpainting using RGBD Diffusion Models}

\author{Jiabao Lei$^1$, \ \ Jiapeng Tang$^2$, \ \ Kui Jia$^{1,3,\dag}$\\
$^1$South China University of Technology \\
$^2$Technical University of Munich, \ \ $^3$Peng Cheng Laboratory
}
\maketitle
\let\thefootnote\relax\footnotetext{
$^\dag$Correspondence to Kui Jia: $<$kuijia@scut.edu.cn$>$.
}

\input{tex/abstract}
\input{tex/intro}

\input{tex/related}

\input{tex/approach}

\input{tex/experiment}

\input{tex/conclusion}

\clearpage
{\small
\bibliographystyle{ieee_fullname}
\bibliography{egbib}
}

\input{tex/inline-supp}

\end{document}

%% file: tex/abstract.tex
\begin{abstract}
We address the challenge of recovering an underlying scene geometry and colors from a sparse set of RGBD view observations. In this work, we present a new solution termed RGBD$^2$ that sequentially generates novel RGBD views along a camera trajectory, and the scene geometry is simply the fusion result of these views. More specifically, we maintain an intermediate surface mesh used for rendering new RGBD views, which subsequently becomes complete by an inpainting network; each rendered RGBD view is later back-projected as a partial surface and is supplemented into the intermediate mesh. The use of intermediate mesh and camera projection helps solve the tough problem of multi-view inconsistency. We practically implement the RGBD inpainting network as a versatile RGBD diffusion model, which is previously used for 2D generative modeling; we make a modification to its reverse diffusion process to enable our use. We evaluate our approach on the task of 3D scene synthesis from sparse RGBD inputs; extensive experiments on the ScanNet dataset demonstrate the superiority of our approach over existing ones.
Project page: \href{https://jblei.site/proj/rgbd-diffusion}{https://jblei.site/proj/rgbd-diffusion}.

\end{abstract}

%% file: tex/intro.tex
\vspace{-0.5cm}
\section{Introduction}
\label{sec:intro}
\vspace{-0.1cm}

Scene synthesis is an essential requirement for many practical applications. The resulting scene representation can be readily utilized in diverse fields, such as virtual reality, augmented reality, computer graphics, and game development. Nevertheless, conventional approaches to scene synthesis usually involve reconstructing scenes (e.g., indoor scenes with varying sizes) by fitting given observations, such as multi-view images or point clouds. The increasing prevalence of RGB/RGBD scanning devices has established multi-view data as a favored input modality, driving and promoting technical advancements in the realm of scene reconstruction from multi-view images.

Neural Radiance Fields (NeRFs) \cite{NeRF} have demonstrated potential in this regard, yet they are not exempt from limitations. NeRFs are designed to reconstruct complete scenes by fitting multi-view images, and they cannot generate or infer missing parts when the input is inevitably incomplete or missing. While recently some studies~\cite{DDP, NeuralRGBD, GRAF, Pi-GAN, GAUDI} have attempted to equip NeRFs with generative and extrapolation capabilities, this functionality relies on a comparatively short representation with limited elements (e.g. typically, the length of a global latent code is much shorter than that of an image: $(F=512) \ll (H\times W=128\times 128=16,384)$) that significantly constrains their capacity to accurately capture fine-grained details in the observed data. Consequently, the effectiveness of these methods has only been established for certain categories of canonical objects, such as faces or cars \cite{Pi-GAN, GRAF}, or relatively small toy scenes \cite{GAUDI}.

\begin{figure}[t]
  \centering
   \includegraphics[width=0.8\linewidth]{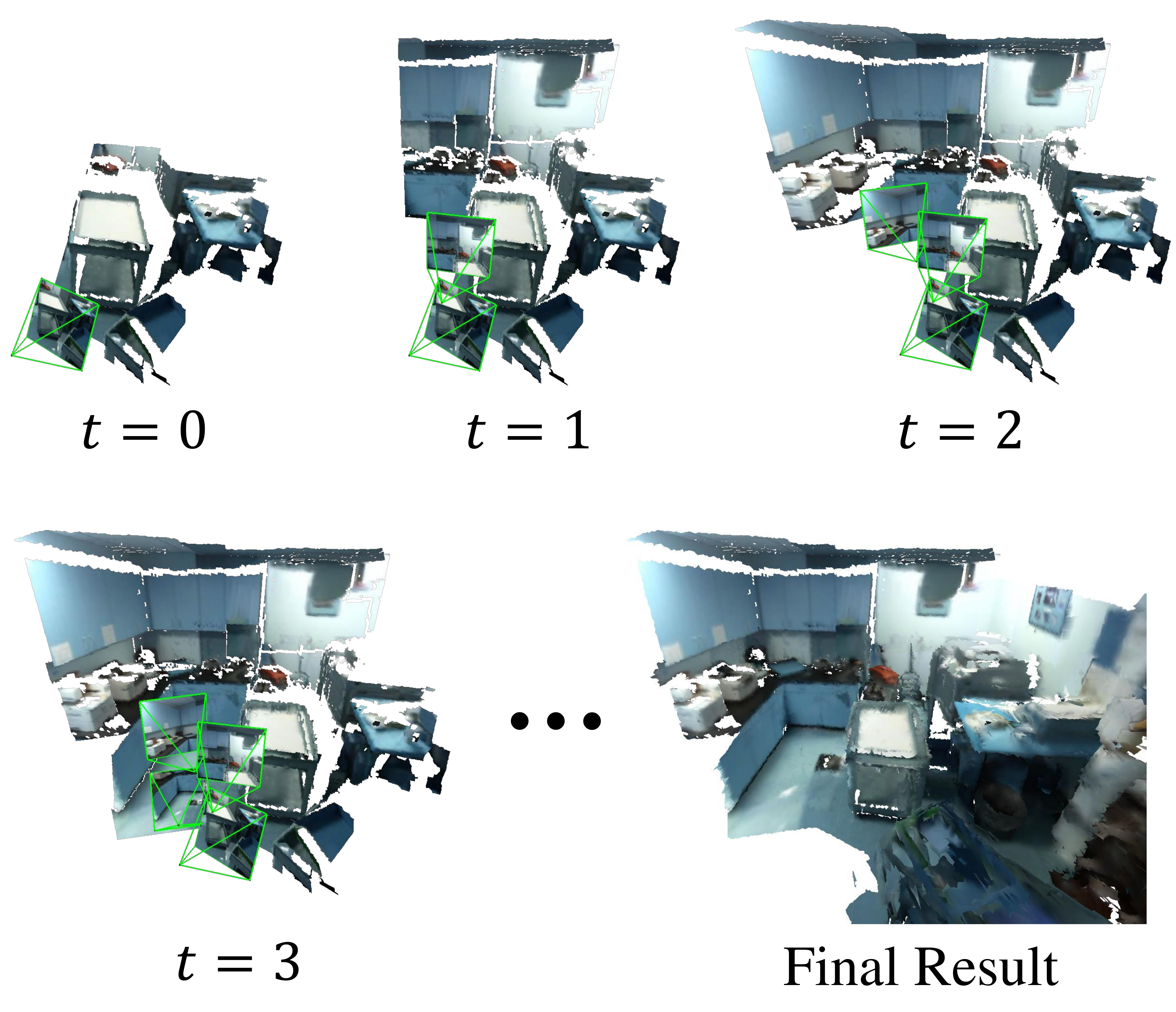}
   \vspace{-0.3cm}
   \caption{\textbf{Illustration of Our Generative Scene Synthesis.} We incrementally reconstruct the scene geometry by inpainting RGBD views as the camera moves in the scene.
   \vspace{-0.5cm}
   }
   \label{fig:teaser}
\end{figure}

We introduce a novel task of generative scene synthesis from sparse RGBD views, which involves learning across multiple scenes to later enable scene synthesis from a sparse set of multi-view RGBD images. This task presents a challenging setting wherein a desired solution should simultaneously (1) preserve observed regions, hallucinate missing parts of the scene, (2) eliminate additional computational costs during inference for each individual test scene, (3) ensure exact 3D consistency, and (4) maintain scalability to scenes with unfixed scales. 

We will elaborate on them in detail as follows.
Firstly, to maximize the preservation of intricate details while simultaneously hallucinating potentially absent parts that may become more pronounced when views are exceedingly sparse, we perform straightforward reconstruction whose details come from images that can describe fine structures using a maximum of $H\times W$ elements (i.e. an image size) in a view completion manner. This is particularly compatible with diffusion models that operate at full image resolution with an inpainting mechanism. 
We also found that RGBD diffusion models greatly simplify the training complexity of a completion model, thanks to their versatile generative ability to inpaint missing RGBD pixels while preserving the integrity of known regions through a convenient training process solely operated on complete RGBD data.
Secondly, our method employs back-projection that requires no optimization, thus eliminating the necessity for test-time training for each individual scene, ultimately leading to a significant enhancement in test-time efficiency.
Thirdly, to ensure consistency among multi-view images, an intermediate mesh representation is utilized as a means of bridging the 2D domain (i.e. multi-view RGBD images) with the 3D domain (i.e. the 3D intermediate mesh) through the aid of camera projection.
Fourthly, to enable our method to handle scenes of indeterminate sizes, we utilize images with freely designated poses as the input representation. Such manner naturally ensures SE(3) equivariance, and thus offers scalability due to the ease with which the range of the generated content can be controlled by simply specifying their camera extrinsic matrices.

Our proposal involves generating multi-view consistent RGBD views along a predetermined camera trajectory, using an intermediate mesh to render novel RGBD images that are subsequently inpainted using a diffusion model, and transforming each RGBD view into a 3D partial mesh via back-projection, and finally merging it with the intermediate scene mesh to produce the final output.
Specifically, our proposed approach initiates by ingesting multiple posed RGBD images as input and utilizing back-projection to construct an intermediate scene mesh. This mesh encompasses color attributes that facilitate the rendering of RGBD images from the representation under arbitrarily specified camera viewpoints. Once a camera pose is selected from the test-time rendering trajectory, the intermediate mesh is rendered to generate a new RGBD image for this pose. Notably, the test-time view typically exhibits only slight overlap with the known cameras, leading to naturally partially rendered RGBD images. To fill the gaps in the incomplete view, we employ an inpainting network implemented as an RGBD diffusion model with minor modifications to its reverse sampling process. The resulting inpainted output is then back-projected into 3D space, forming a partial mesh that complements the entire intermediate scene mesh. We iterate these steps until all test-time camera viewpoints are covered, and the intermediate scene mesh gradually becomes complete during this process. The final output of our pipeline is the mesh outcome acquired from the last step.

Extensive experiments on ScanNet~\cite{ScanNet} dataset demonstrate the superiority of our approach over existing solutions on the task of scene synthesis from sparse RGBD inputs.

%% file: tex/related.tex
\vspace{-0.3cm}
\section{Related Works}
\label{sec:relate}
\vspace{-0.2cm}

In this section, we provide a brief review of the literature related to diffusion models, 3D representations and generative manners, scene synthesis, and view synthesis.

\noindent\textbf{Diffusion Models.}
In recent years, the field of 2D computer vision has experienced a surge of interest in diffusion-based generative models~\cite{DPM, NCSN, DDPM}. These models have prompted the development of image generative modeling approaches, such as GLIDE~\cite{GLIDE}, unCLIP~\cite{unCLIP}, Imagen~\cite{Imagen}, and Latent Diffusion Models~\cite{LDM}, as well as the invention of sampling schedulers~\cite{DDPM, DDIM, PNDM, Euler} and guiding methods~\cite{ClassifierGuided_BeatGAN, ClassifierFree}. Furthermore, these models have been applied to a broad range of image processing tasks, including image inpainting~\cite{RePaint}, image translation~\cite{Palette, SegmentationWithDiffusion, SemanticImageSynthesisviaDiffusionModels, LDM}, video generation~\cite{MCVD, RaMViD, VideoDiffusionModels}, super-resolution~\cite{CascadedDiffusion, SR3, SRDiff}, and image editing~\cite{SDEdit, BlendedDiffusion, DiffusionCLIP}. More recently, some researchers have adapted these techniques from 2D to the 3D domain, as demonstrated by methods such as~\cite{DiffusionPointCloud, PointVoxelDiffusion, DreamFusion, Shape2VecSet}. 
Our approach harnesses such versatility by adopting an iterative denoising strategy like~\cite{RePaint}, and utilizing a masked inpainting technique that operates on the projected RGBD views to synthesize image content.

\noindent\textbf{3D Representations and Generative Manners.}
A variety of representations, including voxels~\cite{3DVoxelGAN2016, O-CNN}, point clouds~\cite{LearningRepresentationsandGenerativeModelsfor3DPointClouds, PointFlow, DiffusionPointCloud, PointVoxelDiffusion, Su-MultiPrototypeLearning}, meshes~\cite{Pixel2Mesh, Atlasnet, PolyGen, SkeletonBridged, SkeletonNet}, implicit surfaces~\cite{OccupancyNetworks, DeepSDF, IM-NET, AnalyticMarching, AnalyticMarchingV2, SAIL-S3, SA-ConvONet, NeuralShapeDeformationPriors, Shape2VecSet, LPDCNet}, multi-view images~\cite{3D-R2N2, DISN, VolumeGuidedProgressiveViewInpainting, InfiniteNature, InfiniteNature-Zero}, and neural radiance fields~\cite{NeRF, GRAF, Pi-GAN, instantNGP, NeuS, HyperNeRF, DS-NeRF, NeuralRGBD, DDP, TANGO}, have been proposed, each with its own unique advantages over the others. This has also motivated researchers to combine them with distinct generative approaches, such as VAEs~\cite{VAE}, GANs~\cite{GAN}, normalizing flows~\cite{NormalizingFlow}, auto-regressive models~\cite{LSTM, AttentionIsAllYouNeed}, and the latest diffusion models~\cite{DDPM}, resulting in an extensive range of applications~\cite{3DVoxelGAN2016, LearningRepresentationsandGenerativeModelsfor3DPointClouds, PointFlow, DiffusionPointCloud, PointVoxelDiffusion, PolyGen, OccupancyNetworks, DeepSDF, GRAF, Pi-GAN}.
However, most existing methods have limitations in their representation capability, such as cubically scaled-up memory consumption, or a fixed number of points, which makes them difficult to apply to scenes of uncertain scales, and poor equivariance, which only allows them to handle canonically-posed objects. In this paper, we address these issues by focusing on the generation of multi-view RGBD images that can capture intricate structures using $H\times W$ pixels created by a diffusion model. This approach reduces memory complexity from $\mathcal{O}(HWD)$ to $\mathcal{O}(HW)$ and increases expressive ability from $\mathcal{O}(F)$ to $\mathcal{O}(HW)$.

\noindent\textbf{Scene Synthesis.} 
In this area, there are two primary research directions. The first pertains to learning configurations, including graphs~\cite{GRAINS, PlanIT, End2endOptSceneLayout, SceneSynthesisHybrid}, top-down views~\cite{FastFlexibleIndoorSceneSynthesis, DeepConvIndoorSceneSynthesis}, and scene composition~\cite{ATISS, SceneFormer}. The second line of research involves direct learning from the appearance of a scene~\cite{NeRF, HyperNeRF, NeuS, instantNGP, DS-NeRF}, thereby obviating the necessity for specialized synthesized datasets required by the aforementioned methods.
Our approach relies solely on RGBD scans, which are readily available from scanning devices, reducing the need for manual annotation. It accurately reconstructs a clean geometry based on the sparse-view input and can effectively hallucinate missing parts, especially when the input views are highly sparse.

\noindent\textbf{View Synthesis.} 
The arrival of NeRF~\cite{NeRF} has significantly advanced the field of view synthesis. While a considerable amount of research has been dedicated to the view synthesis of object-level instances~\cite{EG3D, StyleNeRF, StyleSDF, SceneRepresentationNetworks}, only a few studies have focused on simple scenes~\cite{GeometryFreeViewSynthesis, GenerativeSceneNetworks, GAUDI}. Additionally, some studies have explored techniques for improving performance using sparse inputs~\cite{DDP, RegNeRF}. 
In this paper, we aim to recover scene-level geometry from sparse RGBD images without relying on NeRFs.

%% file: tex/approach.tex
\vspace{-0.2cm}
\section{Preliminary}
\label{sec:ddpm}
\vspace{-0.1cm}
To make our paper self-contained, we provide some preliminary knowledge about DDPM~\cite{DDPM} and DDIM~\cite{DDIM}.

\newcommand\solidrule[1][1cm]{\rule[0.5ex]{#1}{1pt}}
\newcommand\dashedrule{\mbox{%
\solidrule[1mm]\hspace{1mm}
\solidrule[1mm]\hspace{1mm}
\solidrule[1mm]}}

\begin{figure*}[t]
    \centering
    \includegraphics[width=\linewidth]{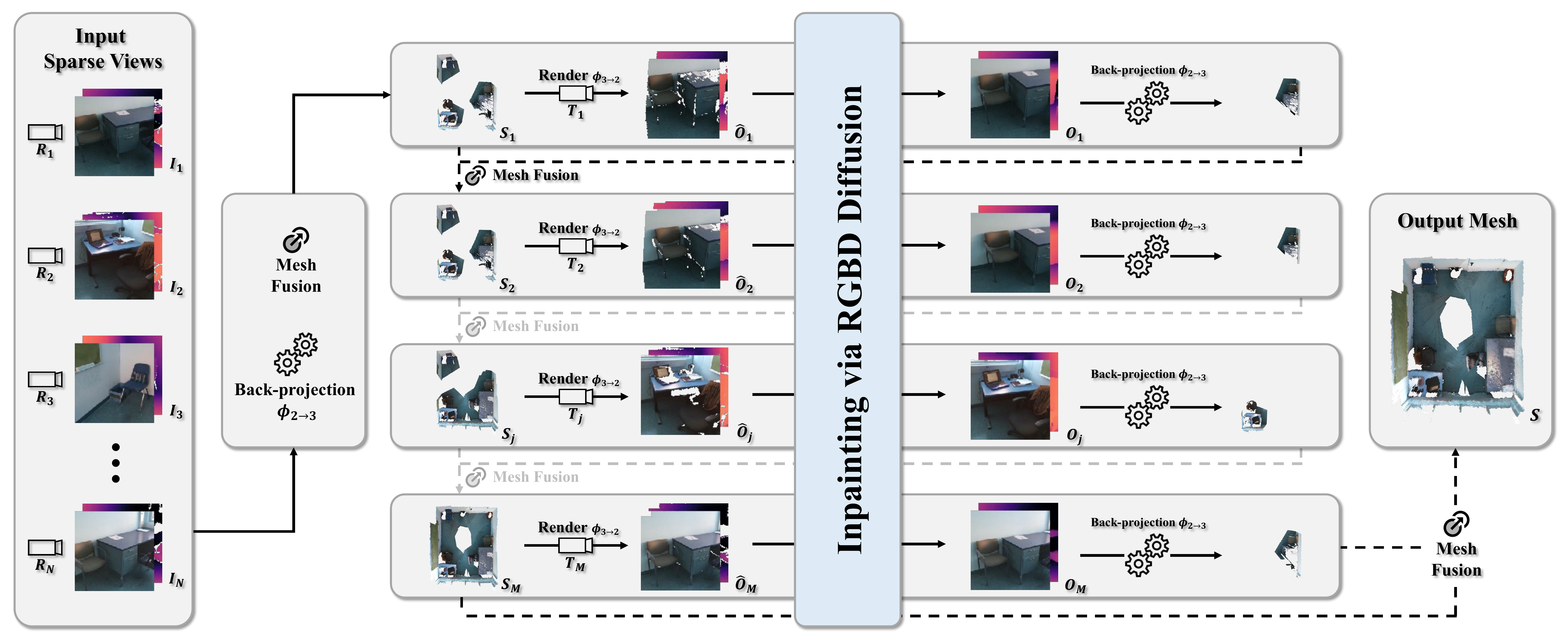}
    \caption{\textbf{3D Scene Synthesis via Incremental View Inpainting.} Given a sparse set of RGBD images $\{ \mathbf{I}_i \}_{i=1}^{N} $ with camera intrinsic $\mathbf{K}$ and extrinsic $\{ \mathbf{R}_i \}_{i=1}^{N}$ matrices, our goal is to generate a coherent 3D scene mesh $\mathcal{S}$ via predicting RGBD frames $\{ \mathbf{O}_j\}_{j=1}^{M}$  along a novel-view trajectory $\{ \mathbf{T}_j\}_{j=1}^{M}$. To achieve this goal, we firstly fuse the inputs of $\{ \mathbf{I}_i \}_{i=1}^{N} $ into an initial mesh $\mathcal{S}_1$, and then render it to obtain an incomplete image $\hat{\mathbf{O}}_1$ that is later inpainted to obtain $\mathbf{O}_1$ using a RGBD diffusion model. After that, $\mathbf{O}_1$ is back-projected and integrated with $\mathcal{S}_1$ to produce a more complete scene, $\mathcal{S}_2$. By iteratively  repeating this process, we can progressively obtain $\mathcal{S}_2, ..., \mathcal{S}_{M}$. Finaly, the fused result $\mathcal{S}_{M+1}$ is the eventual desired output $\mathcal{S}$. Gray dashed lines ``{\protect{\color[gray]{0.75}\dashedrule}}'' denote omitted and unvisualized steps.
    \vspace{-0.5cm}
    }
    \label{fig:scenegen}
\end{figure*}

\noindent \textbf{Definition.}
Given an image $\mathbf{x}_0 \sim q(\mathbf{x}_0)$, the \emph{forward} diffusion process is a Markov chain that sequentially produces noisy images  $\mathbf{x}_1, ..., \mathbf{x}_T$ scheduled by a set of constants $\{\beta_t \in (0, 1) \}_{t=1}^T$ where $T$ is the total number of iterations and $\beta_1 <  \beta_2 < ... < \beta_T$:
\vspace{-0.2cm}
\begin{equation*}
    q(\mathbf{x}_{t} \mid \mathbf{x}_{t-1}) = \mathcal{N}(\mathbf{x}_t; \sqrt{1-\beta_t}\mathbf{x}_{t-1}, \beta_t \mathbf{I})
\end{equation*} 
A nice property of using i.i.d. Gaussian distribution is that we can sample the noisy image $\mathbf{x}_t$ easily based on $\mathbf{x}_0$ in a closed form without computing all the intermediates:
\begin{equation}\label{equa:qxtx0}
    q(\mathbf{x}_{t} \mid \mathbf{x}_0) = \mathcal{N}(\mathbf{x}_t;\sqrt{\bar{\alpha}}_t \mathbf{x}_{0}, (1-\bar{\alpha}_t)\mathbf{I})
\end{equation}
where $\alpha_t=1-\beta_t$ and
$\bar{\alpha}_t=\prod_{s=1}^{t}\alpha_s$.
The \emph{reverse} diffusion process is another Markov chain starting from $\mathbf{x}_T \sim \mathcal{N}(\mathbf{x}_T;\textbf{0}, \textbf{I})$ parameterized by a trainable $\theta$:
\begin{equation}\label{equa:xt-1xt}
    p_\theta(\mathbf{x}_{t-1} \mid \mathbf{x}_{t}) = \mathcal{N}\left(\mathbf{x}_{t-1}; \mathbf{\mu}_\theta(\mathbf{x}_t, t), \mathbf{\Sigma_\theta}(\mathbf{x}_t, t)\right)
\end{equation}
where $\mathbf{\Sigma_\theta}(\mathbf{x}_t, t) = \sigma_t^2 \mathbf{I} = \eta \frac{1 - \bar{\alpha}_{t-1}}{1 - \bar{\alpha}_{t}} \beta_t \mathbf{I}$ \cite{DDIM} is independent of $\theta$ and $\mathbf{x}_t$, and $\mathbf{\mu}_\theta(\mathbf{x}_t, t)$ can be derived from a noise estimator $\mathbf{\epsilon}_\theta(\mathbf{x}_t, t)$ via:
\begin{equation*}
\mathbf{\mu}_\theta(\mathbf{x}_t, \mathbf{x}_o) = \frac{1}{\sqrt{\alpha_t}} \left(\mathbf{x}_t - \frac{1-\alpha_t}{\sqrt{1-\bar{\alpha}_t}} \mathbf{\epsilon}_\theta(\mathbf{x}_t, t)\right)
\end{equation*}
We practically implement the noise estimator $\mathbf{\epsilon}_\theta(\mathbf{x}_t, t)$ as a UNet~\cite{UNet, LDM} parameterized by $\theta$. However, in our case, it is also conditioned on $\hat{\mathbf{x}}_0$ which will be detailed in Sec.~\ref{subsec:rgbdgen}.
\vspace{-0.5cm}
\paragraph{Training.} To train the noise estimator $\mathbf{\epsilon}_\theta(\mathbf{x}_t, t)$, we adopt the simplified training objective~\cite{DDPM}:
\begin{equation*}
\label{equa:loss}
    \min_\theta \mathbb{E}_{t, \mathbf{\epsilon}_t, \mathbf{x}_0 } \left[
   \| \epsilon_\theta(
   \sqrt{\bar{\alpha}_t} \mathbf{x}_{0} + 
   \sqrt{1 - \bar{\alpha}_{t}} \mathbf{\epsilon}_{t}
   , t) - \mathbf{\epsilon}_t \|^2
   \right]
\end{equation*}   
where the time step $t\sim \mathcal{U}\{1, T\}$ is uniformly sampled, $\mathbf{\epsilon}_t \sim \mathcal{N}(\mathbf{0}, \mathbf{I})$ is a standard Gaussian noise, and the image $\mathbf{x}_{0} \sim q(\mathbf{x}_0)$ is randomly drawn from the data distribution.
\vspace{-0.2cm}
\paragraph{Inference.}
We employ a strided DDIM~\cite{DDIM} scheduler to progressively recover the clean image from $\mathbf{x}_T\sim\mathcal{N}(\mathbf{0}, \mathbf{I})$ to $\mathbf{x}_0$ with a subset of $S$ (usually $S \ll T$) steps $\{\tau_i\}_{i=1}^{S}$:
\begin{align*}
\begin{aligned}
    \mathbf{x}_{\tau_{i-1}} \gets 
        \sqrt{\bar{\alpha}_{\tau_{i-1}}} 
        \left(
        \frac{\mathbf{x}_{\tau_{i}} - \sqrt{1 - \bar{\alpha}_{\tau_{i}}}\epsilon_\theta(\mathbf{x}_{\tau_i}, \tau_{i})}{\sqrt{\bar{\alpha}_{\tau_{i}}}}
        \right)
        +
        \\
        \sqrt{1 - \bar{\alpha}_{\tau_{i-1}} - \sigma_{\tau_{i}}^2} \cdot \epsilon_{\theta}(\mathbf{x}_{\tau_{i}}, \tau_i)
        +
        \sigma_{\tau_{i}}^2 \epsilon_{\tau_{i}}
\end{aligned}
\vspace{-0.3cm}
\end{align*}
where $\epsilon_{\tau_{i}} \sim \mathcal{N}(\mathbf{0}, \mathbf{I})$ is a standard Gaussian noise. It is interesting that setting $\eta=0$ implies $\forall i, \ \sigma_{\tau_{i}} \equiv 0$, making the inference process deterministic. Nevertheless, diversity can still be achieved by choosing a different $\mathbf{x}_T$ from $\mathcal{N}(\mathbf{0}, \mathbf{I})$.


\vspace{-0.3cm}
\section{Approach}
\label{sec:app}
\vspace{-0.1cm}
In this work, we introduce a new scene synthesis approach that relies on the sequential generation of RGBD frames using a trained RGBD diffusion model conditioned on views rendered under a test-time camera trajectory from a sparse set of input RGBD images.
Our approach leverages the generative ability of diffusion models to synthesize missing visual appearance (color) and geometry details (depth) while rigorously preserving visible regions by image inpainting.
Such an incremental view inpainting process is interleaved with the back-projection, mesh fusion, and mesh rendering procedures to achieve global 3D consistency among distinct temporal frames.

This section is organized as follow.
In Sec.~\ref{subsec:scenegen}, we will start by introducing the overall framework of repeatedly converting the RGBD image into a partial 3D mesh and performing RGBD inpainting for rendered views in a progressive manner. 
Later in Sec.~\ref{subsec:rgbdgen}, we will elaborate more on the details of the conditional RGBD diffusion model for rendered view inpainting based on a trained DDPM~\cite{DDPM}.

\vspace{-0.2cm}
\subsection{Incremental View Inpainting Fashion}
\label{subsec:scenegen}
\vspace{-0.1cm}

As shown in Figure~\ref{fig:scenegen},  given a sparse set of $N$ RGBD views $\{ \mathbf{I}_i \}_{i=1}^{N} $ with their associated camera intrinsic $\mathbf{K}$ and extrinsic $\{ \mathbf{R}_i \}_{i=1}^{N} $ matrices, and additionally a camera trajectory composed of $M$ viewpoints $\{ \mathbf{T}_j\}_{j=1}^{M}$ with the same intrinsic matrix, the essence of our method is to progressively synthesize RGBD images $\mathbf{O}_j$ at each specified novel view $\mathbf{T}_j$. The generated novel-view frames should be consistent with $\{ \mathbf{I}_i \}_{i=1}^{N}$ in both geometry and appearance. The final output should be a consistent 3D colored mesh $\mathcal{S}$ converted from the back-projection result $ 
\left(  \cup_{j=1}^{M}\phi_{2\rightarrow3}(\mathbf{O}_j) \right) \cup 
\left(  \cup_{i=1}^{N}\phi_{2\rightarrow3}(\mathbf{I}_i) \right)$. 
Please also refer to Algorithm~\ref{alg:autogen} for a detailed description of the procedure.
\begin{algorithm}[hbt!]
    \caption{Incremental RGBD View Inpainting}\label{alg:autogen}
    \begin{algorithmic}
        \Require RGBD images $\{ \mathbf{I}_i \}_{i=1}^{N}$ with camera extrinsics $\{ \mathbf{R}_i \}_{i=1}^{N}$, a novel trajectory of M extrinsics $\{ \mathbf{T}_j \}_{j=1}^{M}$.
        \State $\mathcal{S}_1 \gets \cup_{i=1}^{N} \phi_{2\rightarrow3}(\mathbf{I}_i)$
        \For{$j=1$ to $M$}
            \State $\hat{\mathbf{O}}_j \gets \phi_{3\rightarrow2}(\mathbf{T}_j)$  \Comment{novel-view rendering}
            \State $\mathbf{O}_j \gets f(\hat{\mathbf{O}}_j)$  \Comment{RGBD inpainting}
            \State $\mathcal{S}_{j+1} \gets \mathcal{S}_j \cup \phi_{2\rightarrow3}(\mathbf{O}_j) $ \Comment{mesh fusion}
        \EndFor \\
        \Return $ \mathcal{S}_{M+1} $
    \end{algorithmic}
\end{algorithm}

\noindent \textbf{Rendering and Back-projection.\hyperref[link1]{$^1$}}\footnote{\label{link1}$^1$ To simplify notations, we disregard certain arguments in operators $\phi_{3\rightarrow2}$ and $\phi_{2\rightarrow3}$, such as the camera intrinsic matrix $\mathbf{K}$.
}
The operator $\phi_{3\rightarrow2}$ is implemented as mesh rasterization, which allows for the rendering of a partial RGBD image $\hat{\mathbf{O}}_j$ from a mesh $\mathcal{S}_j$. This approach offers the advantage of producing a clean visibility mask $\mathbf{m}_j$, which is not possible with NeRFs~\cite{NeRF}. The back-projection operator $\phi_{2\rightarrow3}$ is responsible for the conversion from a depth map into a point cloud, where the connectivity between points is inherited from the connectivity of the 2D pixel grid. Furthermore, mesh faces that are either in close proximity to the viewpoint or exhibit slender characteristics are filtered out to ensure accuracy.

\noindent \textbf{Challenges and Solutions.}
To circumvent the limitations posed by potential 3D inconsistency in both geometry and appearance, as well as the challenge of solely handling specific canonically-posed scenes, we propose several strategies to address these issues.
Firstly, our method for synthesizing novel views combines rendering (mesh rasterization) and inpainting techniques, and interleaves the view synthesis process with online RGBD fusion via back-projection and mesh combination. The use of perspective camera projection ensures strict adherence to 3D constraints, resulting in visually consistent and accurate synthesized views.
In concrete terms, we begin by rendering the mesh $\mathcal{S}_j$ under view $\mathbf{T}_j$ using a rendering operation $\phi_{3\rightarrow2}(\mathbf{T}_j)$. This process yields an incomplete RGBD image $\hat{\mathbf{O}}_j$ with missing regions, which is subsequently inpainted using a diffusion model described in detail in Sec.~\ref{subsec:rgbdgen}, resulting in a complete image $\mathbf{O}_j$. Once image $\mathbf{O}_j$ has been generated for view $\mathbf{T}_j$, it can be fused into a 3D mesh via $\mathcal{S}_{j+1} = \phi_{2\rightarrow3}(\mathbf{O}_j) \cup \mathcal{S}_{j}$ using a back-projection operator $\phi_{2\rightarrow3}$.
Secondly, our solution reduces the learning difficulty and can handle noncanonical scenes by decomposing the 3D scene as a Markov chain of temporal RGBD images rendered from arbitrarily specified novel viewpoints. 
The applicability of handling scenes with arbitrary scaling, movement, and posing is attributed to the utilization of two SE(3) equivariant operators, namely $\phi_{3\rightarrow2}$ and $\phi_{2\rightarrow3}$, as well as the independence of absolute coordinates.
Moreover, the presence of redundant information in adjacent frames, combined with our suggested decomposition rule, facilitates the minimization of learning complexity in an auto-regressive manner. In accordance with the Markov chain decomposition principle, the distribution of the scene can be expressed as the joint distribution of view frames:
\begin{equation*}
\vspace{-0.1cm}
    p(\mathcal{S}) = \prod_j p(\mathbf{O}_j \mid \{\mathbf{O}_{s}\}_{s=1}^{j-1},  \ \{\mathbf{I}_{s}\}_{s=1}^{N})
\vspace{-0.1cm}
\end{equation*}
where the prediction of $\mathbf{O}_j$ is based on the fusion result of all the previously known frames $\{\mathbf{O}_{s}\}_{s=1}^{j-1} \cup \{\mathbf{I}_{s}\}_{s=1}^{N}$.

\subsection{RGBD Diffusion for Rendered View Inpainting}
\label{subsec:rgbdgen}

In this section, we describe the implementation details of the way to employ diffusion models to inpaint the missing regions of the RGBD image $\hat{\mathbf{O}}_j$ with a binary mask $\mathbf{m}_j$ obtained by rendering visibility, where the value of 1 is assigned to the corresponding ray that intersects with the geometry surface, while 0 is assigned to all other cases, both of which are rendered by projecting $\mathcal{S}_j$ under a novel viewpoint $\mathbf{T}_j$.
It is noteworthy that only those pixels located where $\mathbf{m}_j=0$ in $\hat{\mathbf{O}}_j$ are considered invalid, and therefore, are entirely filled with zeros (i.e., $\hat{\mathbf{O}}_j \odot \mathbf{m}_j \equiv \hat{\mathbf{O}}_j$).

\vspace{-0.1cm}
\begin{figure}[!htp]
    \centering
    \includegraphics[width=\linewidth]{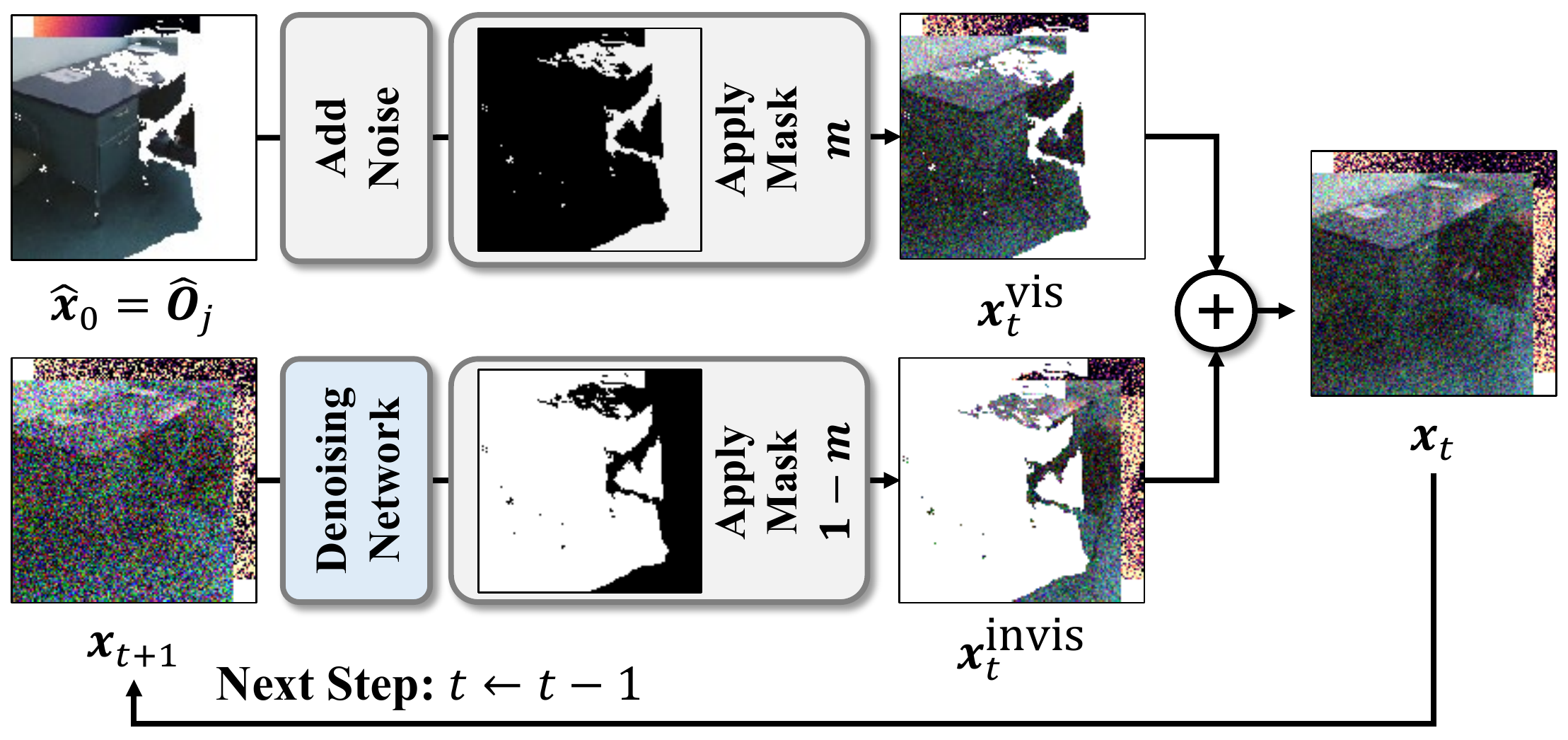}
    \caption{\textbf{RGBD Diffusion for Rendered View Inpainting.}
    In each iteration, we utilize a forward diffusion step to sample the visible region $\mathbf{x}_{t}^{\textrm{vis}}$ (\emph{top}) from the incomplete input $\hat{\mathbf{x}}_0$, while employing a denoising diffusion step to obtain the complementary content $\mathbf{x}_{t}^{\textrm{invis}}$ (\emph{bottom}) from the output $\mathbf{x}_{t+1}$ generated in the previous iteration $t + 1$.
    \vspace{-0.2cm}
    }
\end{figure}

\noindent \textbf{A Single Reverse Diffusion Step for $\mathbf{T}_j$ at Time $t$.\hyperref[link2]{$^2$}}\footnote{\label{link2}$^2$ For notational simplicity, we ignore the novel camera subscript $j$ in the following part of this Sec.~\ref{subsec:rgbdgen} since all notations we consider here are derived from the same camera $\mathbf{T}_j$.}
We initialize the clean diffusion sample $\hat{\mathbf{x}}_0$ at time $t=0$ as $\hat{\mathbf{O}}_j$, and those noisy samples at time $0 < t \le T$ are denoted as $\mathbf{x}_t$.
To ensure 3D consistency, we seek to predict the invisible region  $ \mathbf{x}_0^{\textrm{invis}}$ while preserving the visible portion $\mathbf{x}_0^{\textrm{vis}}$ unaltered.
We follow~\cite{RePaint} and modify the standard denoising process to meet our needs.
At each reverse diffusion step $t$, we use the forward diffusion process defined by the Eqn.~(\ref{equa:qxtx0}) to sample a visible region $ \mathbf{x}_{t}^{\textrm{vis}}$ masked by $\mathbf{m}$, and the reverse diffusion step defined by Eqn.~(\ref{equa:xt-1xt}) to sample a full part from $\mathbf{x}_{t+1}$ which is later masked by $1-\mathbf{m}$ to form an invisible region $ \mathbf{x}_{t}^{\textrm{invis}}$.
\begin{equation*}
\mathbf{x}_{t}^\textrm{{vis}} = \sqrt{\bar{\alpha}_{t}} \hat{\mathbf{x}}_{0} + (1-\bar{\alpha}_{t}) \mathbf{\epsilon}_{t}^{\textrm{vis}} \odot \mathbf{m} 
\end{equation*}
\vspace{-0.5cm}
\begin{equation*}
\mathbf{x}_{t}^\textrm{{invis}} = \left(
\mathbf{\mu}_\theta \left(\mathbf{x}_t, \hat{\mathbf{x}}_t, t\right) + \sigma_t \mathbf{\epsilon}_{t}^{\textrm{invis}} \right) \odot (1 - \mathbf{m})
\end{equation*}
where $\mathbf{\epsilon}_{t}^{\textrm{vis}}, \mathbf{\epsilon}_{t}^{\textrm{invis}} \sim \mathcal{N}(\mathbf{0}, \mathbf{I})$. 
The noisy image $\mathbf{x}_t$ can be simply calculated as the sum of $\mathbf{x}_{t}^{\textrm{vis}}$ and $\mathbf{x}_{t}^{\textrm{invis}}$, as expressed by $\mathbf{x}_t = \mathbf{x}_t^\textrm{{vis}} + \mathbf{x}_t^{\textrm{invis}}$.
\vspace{-0.3cm}

\vspace{-0.1cm}
\noindent \paragraph{Diffusion Network.}
We implement the diffusion model as a UNet \cite{UNet, LDM} conditioned on the observed region $\mathbf{x}_0$. We input the concatenation of $\mathbf{x}_t$ and $\hat{\mathbf{x}}_0$ into the network.

\vspace{-0.5cm}
\noindent \paragraph{Classifier-free Guidance.}
To further enhance the controllability of the generation process, we introduce a classifier-free guidance~\cite{ClassifierFree} mechanism.
Specifically, we train a unified network $\epsilon_\theta$ comprising of an unconditional model $\epsilon_\theta(\mathbf{x}_t, \mathbf{c}, t)$, where the shared variable $\mathbf{c}$ is incorporated, and a conditional model $\epsilon_\theta(\mathbf{x}_t, \hat{\mathbf{x}}_0, t)$.
In this way, the predicted noise $\Tilde{\epsilon}_\theta$ can be recomputed as follows:
\begin{equation*}
    \Tilde{\epsilon}_\theta(\mathbf{x}_t, \hat{\mathbf{x}}_0, t) = 
    \epsilon_\theta(\mathbf{x}_t, \mathbf{c}, t) + 
    \beta \times [
    \epsilon_\theta(\mathbf{x}_t, \hat{\mathbf{x}}_0, t)
    -
    \epsilon_\theta(\mathbf{x}_t, \mathbf{c}, t)
    ]
\end{equation*}
where $\beta \ge 0$ is the guidance factor, being responsible for the trade-off between sampling quality and diversity~\cite{ClassifierFree}.

%% file: tex/experiment.tex
\vspace{-0.2cm}
\section{Experiments}
\label{sec:exper}
\vspace{-0.2cm}

\noindent\textbf{Dataset.} 
We conducted experiments on the ScanNet-V2~\cite{ScanNet} dataset, which was pre-processed by removing redundant frames~\cite{NeuralRecon}. 
For training, we used the first $1,293$ scenes, while for metric evaluation, we randomly selected 18 scenes with over 50 views each from the remaining as our test set. 
We also evaluated under various sparsity settings (5\%, 10\%, 20\%, and 50\%) by uniformly down-sampling views.

\noindent\textbf{Comparison.} 
We compared against the neural graphics primitive (NGP)~\cite{instantNGP}, which has demonstrated impressive performance in scene modeling with high efficiency. 
To enhance its geometric quality, we incorporated a depth supervision (DS) loss~\cite{DS-NeRF} to build an improved variant called DS-NGP~\cite{DS-NeRF, instantNGP}.
We also compare against Neural RGBD (N-RGBD)~\cite{NeuralRGBD}, which recovers implicit surfaces from RGBD scans, and Dense Depth Prior (DDP)~\cite{DDP}, which learns a NeRF utilizing view completion from sparse RGBD views.

\noindent\textbf{Evaluation Metrics.} 
For assessing the visual quality of RGB images, we adopted the peak signal-to-noise ratio (PSNR), structural similarity index measure (SSIM), and learned perceptual image patch similarity (LPIPS)~\cite{LPIPS} that is based on the AlexNet~\cite{AlexNet} backbone. 
To evaluate the geometry quality, we computed the mean squared error (MSE) on depth maps, and sample $10,000$ points uniformly on meshes constructed via back-projection to evaluate the chamfer distance (CD) and completeness (Comp.) with a threshold of 0.1m. 
We also measured the computational time required to execute different stages of the method. 
For LPIPS, MSE, and CD, the lower the better; for PSNR, SSIM, and Comp., the higher the better.
All reported metrics are averaged across the test scenes.

\noindent\textbf{Implementation Details.} 
Our model has 157M parameters and was trained for three days on 7 NVIDIA 3090-Ti GPUs, using a batch size of 280. The learning rate was initialized at $1\times10^{-4}$ and reduced to $1\times10^{-6}$ over a period of 300 epochs, utilizing a cosine annealing strategy.
The image resolution is $128\times 128$ with a rendering chunk size of 7.

\vspace{-0.1cm}
\subsection{Ablation Studies}
\label{subsec:experscenegen}
\vspace{-0.1cm}

We conducted ablation studies to validate the effectiveness of each component in our proposal. Results concluded here are nontrivial and a bit nuanced.

\begin{figure*}[ht]
    \centering
    \scalebox{0.9} {
        \begin{minipage}[t]{0.06\linewidth}%
            \scalebox{0.8}{\rotatebox{90}{%
            \begin{tabular}[t]{>{\centering\arraybackslash}p{1.86cm}
                               >{\centering\arraybackslash}p{1.86cm}
                               >{\centering\arraybackslash}p{1.86cm}
                               >{\centering\arraybackslash}p{1.96cm}
                               >{\centering\arraybackslash}p{1.10cm}
                               }
                \shortstack{$\beta=5$ \\ (d)} & 
                \shortstack{$\beta=1$ \\ (c)} &
                \shortstack{w/o inpa. \\ $\beta=1$ \\ (b)} &
                \shortstack{$\beta=0$ \\ (a)} &
                Input
            \end{tabular}
            }}
        \end{minipage}
        \begin{minipage}[t]{0.94\linewidth}
            \centering
            \includegraphics[width=\linewidth]{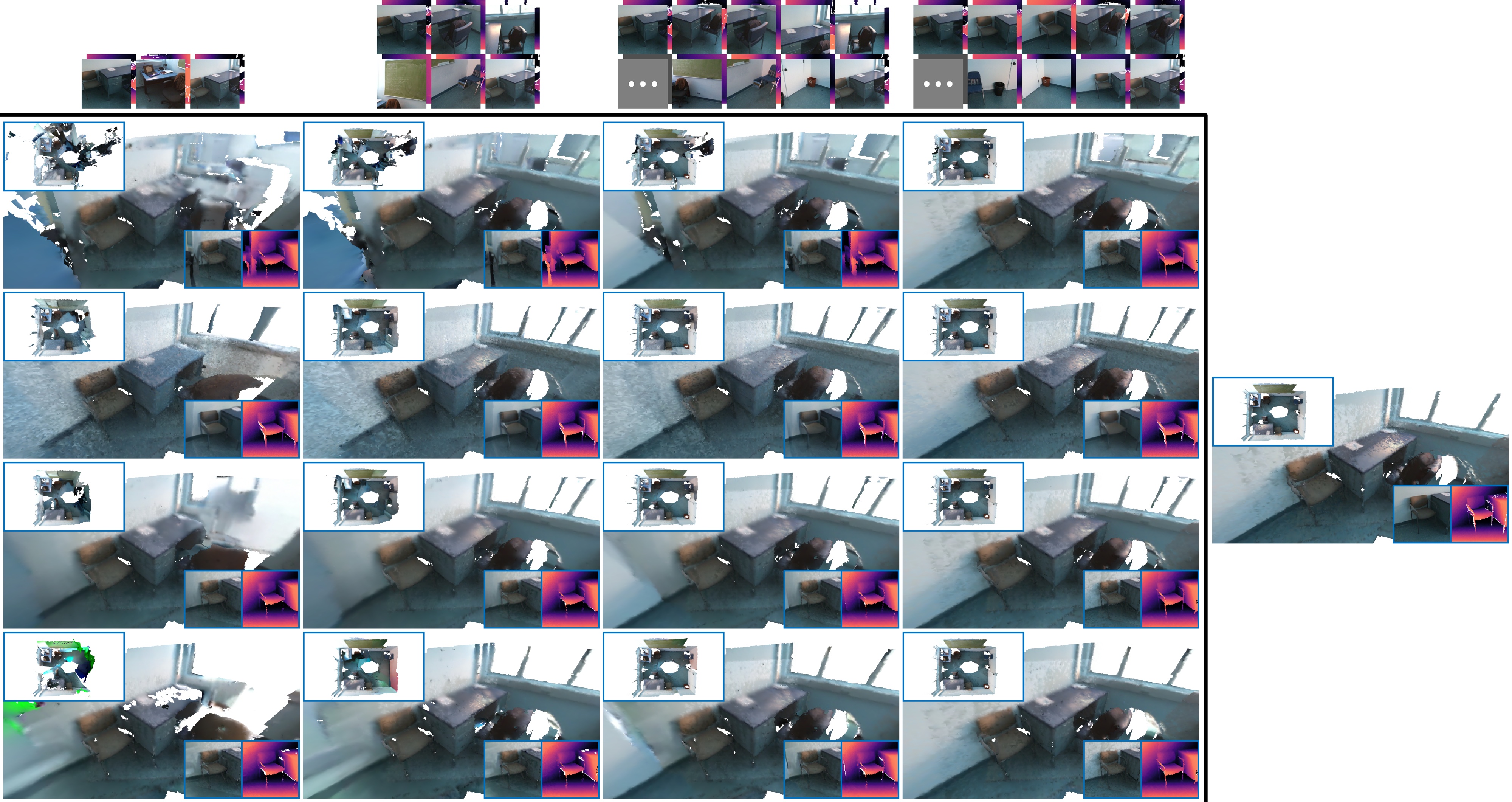}
            \begin{tabular}[t]{>{\centering\arraybackslash}p{2.9cm}
                                   >{\centering\arraybackslash}p{2.9cm}
                                   >{\centering\arraybackslash}p{2.8cm}
                                   >{\centering\arraybackslash}p{2.8cm}
                                   >{\centering\arraybackslash}p{3.2cm}
                                   }
                    (i) 5\%  &
                    (i) 10\%  &
                    (i) 20\%  &
                    (i) 50\% &
                    GT
                \end{tabular}
        \end{minipage}
    }
    \vspace{-0.3cm}
    \caption{\textbf{Qualitative results of ablation studies on the task of scene synthesis from sparse RGBD inputs. }
    ``(a)'', ``(b)'' and ``(c)'' correspond to the second (w/o conditioning, $\beta=0$), third (w/o inpainting, $\beta=1$) and fourth (w/ all, $\beta=1$) rows of Table~\ref{table:abla_compo}, and ``(d)'' corresponds to the last (w/ all, $\beta=5$) row of Table~\ref{table:abla_scale}. 
    Each image in the grid consists of four sub-images, with two located at the lower-right corner displaying rendered RGBD images generated from either NeRFs (others) or meshes (ours), and the other two positioned at the top-left corner and underneath, respectively, showcasing the back-projected triangular meshes captured from a close-up perspective.
    It is clear that the third row basically shows the most favorable appearance. 
    \vspace{-0.3cm}
    }
    \label{fig:abla_1}
\end{figure*}

\noindent\textbf{Effects of Different Ingredients.} 
We examined the anticipated efficacy of the proposed conditioning and inpainting components in our approach. 
Table~\ref{table:abla_compo} displays the numerical results, and Figure~\ref{fig:abla_1}-(a-c) presents the visualizations. 
The combined use of conditioning and inpainting yields superior visual performance compared to the ground truth. Nevertheless, when the model is conditioned, the impact of inpainting on geometric quality becomes less crucial. 
Moreover, as more views are provided, the stochastic generation process becomes increasingly deterministic, resulting in structures that more closely resemble the ground truth.


\noindent\textbf{Effects of Guidance Scale.}
The guidance factor $\beta$ significantly affects the conditioning effect on the results. 
To investigate the optimal $\beta$ under various settings, we conducted experiments with $\beta$ chosen from 0.0, 0.5, 1.0, 2.0, and 5.0, respectively, and evaluated their performance. Quantitative results are presented in Table~\ref{table:abla_scale}. 
It is interesting that only a suitable value of $\beta$ (1.0 or 2.0) yields optimal performance, while smaller or larger values of $\beta$ result in underperformance.
Surprisingly, for geometric recovery (MSE and CD), the optimal value of $\beta$ appears to increase as the percentage gets larger. 
For a scene with a percentage $\ge 50\%$, the best value of $\beta$ is amazingly greater than 5.0. However, such a large percentage is not optimal for visual appearance. 
This is because using a larger value of $\beta$ at a low percentage can cause the generated results to deviate unexpectedly from the ground truth.
Our visualization, presented in Figure~\ref{fig:abla_1}-(a, c, d), indicates that using an unconditional model ($\beta=0$) leads to undesired and bizarre geometric structures since the network fails to understand the context provided by known views. However, excessively large values of $\beta$ also oversaturate the color (e.g. row ``(d)'' at $5\%$), making the visual appearance unrealistic.


\noindent\textbf{Effects of Randomness and Trajectories.}
\begin{figure}[ht]
    \centering
    \includegraphics[width=\linewidth]{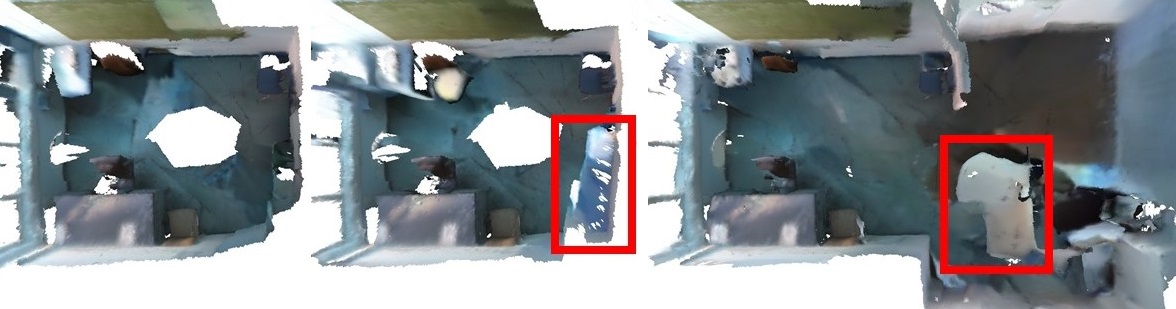}
    \begin{tabular}[t]{>{\centering\arraybackslash}p{1.7cm}
                       >{\centering\arraybackslash}p{1.9cm}
                       >{\centering\arraybackslash}p{3.5cm}
                    }
        \vspace{-0.4cm} (a) original &
        \vspace{-0.4cm} (b) seed &
        \vspace{-0.4cm} (c) trajectory
    \end{tabular}
    \vspace{-0.6cm}
    \caption{\textbf{Diverse results obtained by changing the (b) random seed and (c) camera trajectory based on (a) the original scene.}
    \vspace{-0.9cm}
    }
    \label{fig:diversity}
\end{figure}
We investigated the impact of employing different random seeds and camera trajectories on the results. 
Such obtained meshes are visualized in Figure~\ref{fig:diversity}.
As diffusion models inherently include randomness in their reverse sampling process, by switching to a different random seed, we can obtain another outcome. 
Moreover, the camera trajectory can affect the rough shape of the scene. 
It is promising that our approach can yield controllable and editable results, as it is capable of generating various and appealing outcomes by simply altering these two factors.


\vspace{-0.1cm}
\subsection{Generative Scene Synthesis}
\begin{figure*}[ht]
    \centering
    \scalebox{0.9}{
        \centering
        \begin{minipage}[t]{0.05\linewidth}%
            \scalebox{0.8}{\rotatebox{90}{%
            \begin{tabular}[t]{>{\centering\arraybackslash}p{1.83cm}
                               >{\centering\arraybackslash}p{1.83cm}
                               >{\centering\arraybackslash}p{1.83cm}
                               >{\centering\arraybackslash}p{1.93cm}
                               >{\centering\arraybackslash}p{1.10cm}
                               }
                Ours & 
                \shortstack{\cite{NeuralRGBD} \\ N-RGBD } &
                \shortstack{\cite{DDP} \\ DDP } &
                \shortstack{\cite{instantNGP, DS-NeRF} \\ DS-NGP} &
                Input
            \end{tabular}
            }}
        \end{minipage}
        \begin{minipage}[t]{0.95\linewidth}%
            \centering
            \includegraphics[width=\linewidth]{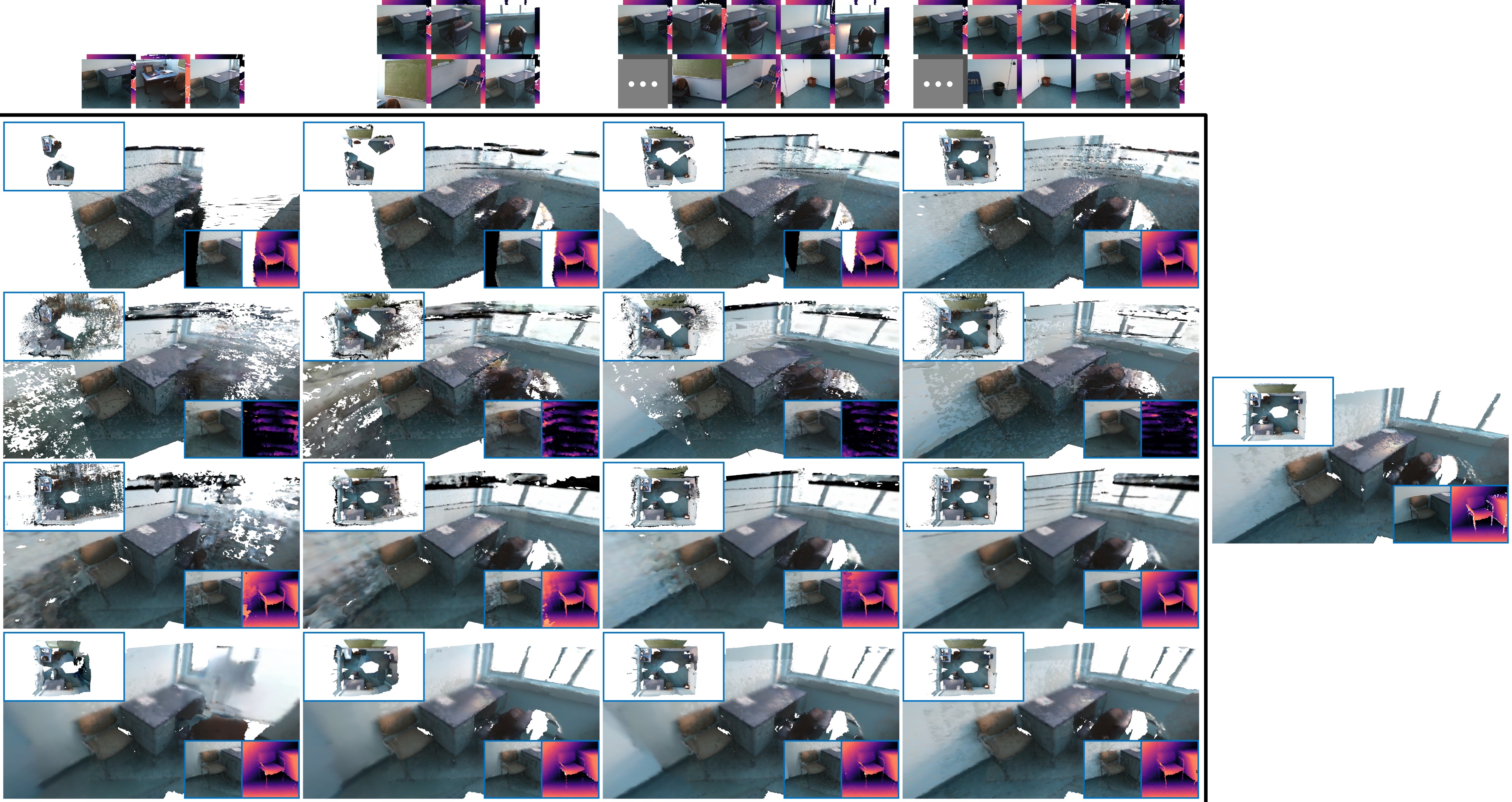}
        \end{minipage}
    }
    \scalebox{0.9}{
        \centering
        \begin{minipage}[t]{0.05\linewidth}%
            \scalebox{0.8}{\rotatebox{90}{%
            \begin{tabular}[t]{>{\centering\arraybackslash}p{1.83cm}
                               >{\centering\arraybackslash}p{1.83cm}
                               >{\centering\arraybackslash}p{1.83cm}
                               >{\centering\arraybackslash}p{1.93cm}
                               >{\centering\arraybackslash}p{1.10cm}
                               }
                Ours & 
                \shortstack{\cite{NeuralRGBD} \\ N-RGBD  } &
                \shortstack{\cite{DDP} \\ DDP  } &
                \shortstack{\cite{instantNGP, DS-NeRF} \\ DS-NGP} &
                Input
            \end{tabular}
            }}
        \end{minipage}
        \begin{minipage}[t]{0.95\linewidth}%
            \centering
            \includegraphics[width=\linewidth]{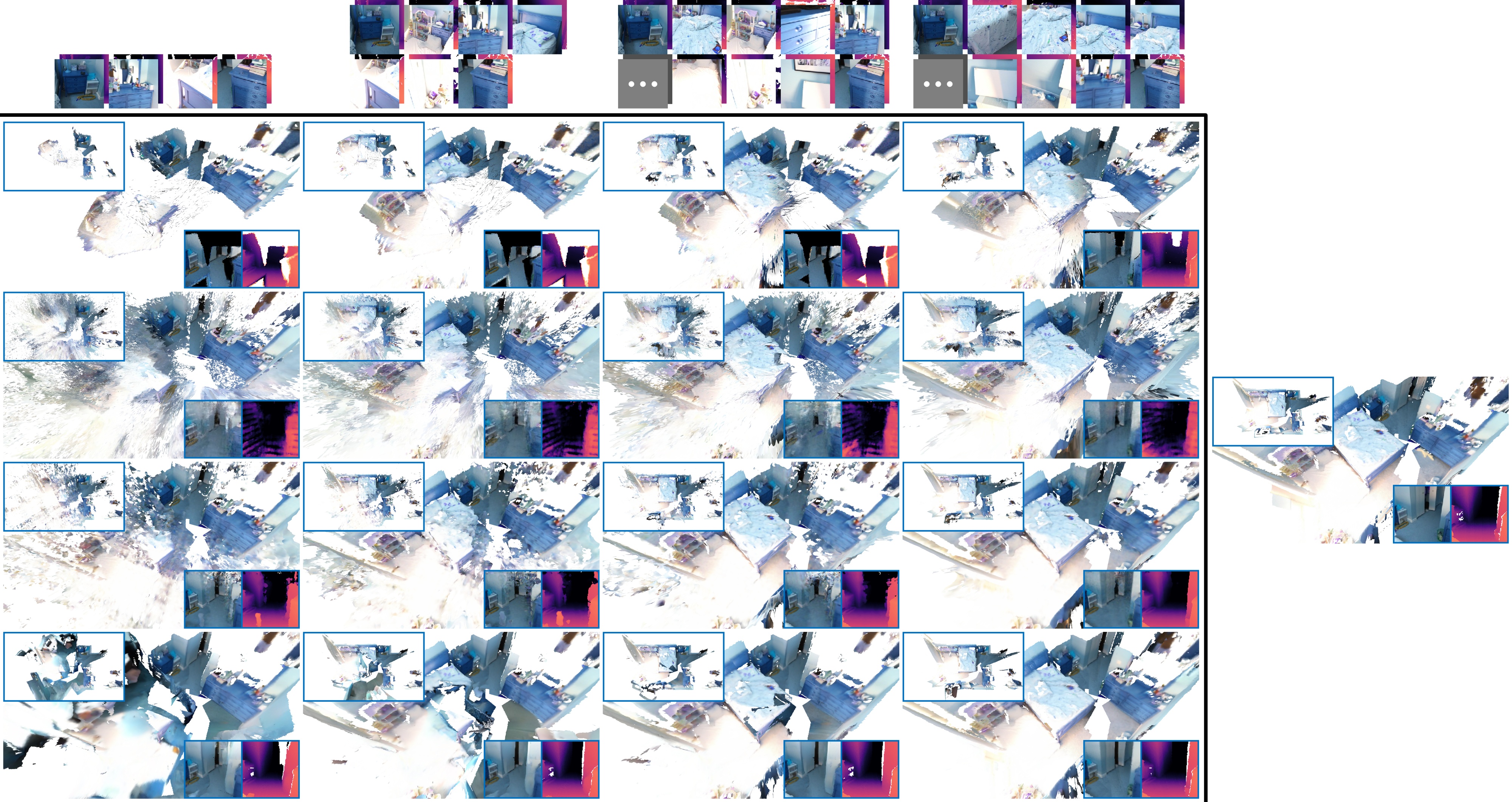}
            \begin{tabular}[t]{>{\centering\arraybackslash}p{2.9cm}
                               >{\centering\arraybackslash}p{2.9cm}
                               >{\centering\arraybackslash}p{2.9cm}
                               >{\centering\arraybackslash}p{2.9cm}
                               >{\centering\arraybackslash}p{3.2cm}
                               }
                (i) 5\%  &
                (i) 10\%  &
                (i) 20\%  &
                (i) 50\% &
                GT
            \end{tabular}
        \end{minipage}
    }
    \vspace{-0.3cm}
    \caption{\textbf{Qualitative comparison results with DS-NGP~\cite{instantNGP, DS-NeRF}, DDP~\cite{DDP}, N-RGBD~\cite{NeuralRGBD} on the task of scene synthesis from sparse RGBD inputs.}
        Each image in the grid consists of four sub-images, with two located at the lower-right corner displaying rendered RGBD images generated from either NeRFs (others) or meshes (ours), and the other two positioned at the top-left corner and underneath, respectively, showcasing the back-projected triangular meshes captured from a close-up perspective.
        Our approach produces images that exhibit a close resemblance to the ground truth (GT), while comparative methods may result in inferior outcomes, especially when dealing with excessively sparse input views and outward-looking cameras.
        \vspace{-0.3cm}
    }
    \label{fig:novel_view}
\end{figure*}

\vspace{-0.1cm}
We evaluated the performance of our method against other similar approaches on this task through extensive experiments under various sparsity settings. 
Table~\ref{table:time} presents the time required by different methods across various stages. 
Our approach stands out for its efficiency, as it eliminates the need for additional optimization on individual test scenes during inference.
\begin{table}[ht]
    \centering
    \vspace{-0.2cm}
    \resizebox{\linewidth}{!} {
        \begin{tabular}[t]{*{6}{c}}
             \toprule
             Time                      & Backend & Repre. & Training  &  Optimization &  Rendering \\
             \midrule
             (DS-)NGP~\cite{instantNGP, DS-NeRF}  & CUDA C++ & NeRF & ---  & $\sim3$ min.   & $\sim0.05$ sec.        \\
             N-RGBD~\cite{NeuralRGBD}  & PyTorch & NeRF & ---  & $\sim2$ hours               & $\sim2$ sec.            \\
             DDP~\cite{DDP}            & PyTorch & NeRF & $\sim1$ day       & $\sim1$ hour        & $\sim1$ sec.            \\
             Ours                      & PyTorch & Mesh & $\sim3$ days       & ---        & $\sim3$ sec.    \\
             \bottomrule
        \end{tabular}
    }
    \vspace{-0.3cm}
    \caption{
    \textbf{Efficiency Comparison of Different Methods.}
    It details their distinct implementation backends, the underlying representation (Repre.) methods, and the time required for \emph{training} on multiple training scenes, \emph{optimization} on a single test scene, and \emph{rendering} a single view after per-scene optimization.
    \vspace{-0.3cm}
    }
    \label{table:time}
\end{table}
Quantitative results are shown in Table~\ref{table:novel_view}.
We observe that our approach exhibits a clear advantage in visual metrics over the others when the provided views are extremely sparse (5\%). Interestingly, our method consistently achieves the best performance in all the geometry-related metrics.
Qualitative results are presented in Figure~\ref{fig:novel_view}.
DS-NGP~\cite{instantNGP, DS-NeRF} and DDP~\cite{DDP} struggle to accurately recover geometry due to their inability to hallucinate and extrapolate missing regions. 
In comparison, N-RGBD~\cite{NeuralRGBD} can achieve better surface completion results by learning and extrapolating neural implicit surfaces.
However, their performance significantly degrades when input views are extraordinarily sparse (5\%). 
In contrast, our method consistently exhibits plausible appearance that closely resembles the ground truth, particularly in scenarios with sparse-view inputs.

\begin{table*}[!ht]
    \centering
    \resizebox{\textwidth}{!} {
        \begin{tabular}[t]{*{26}{c}}
             \toprule
               \multicolumn{2}{c}{\multirow{2}{*}{\vspace{-0.2cm}Factors}} & 
               \multicolumn{12}{c}{Visual} & 
               \multicolumn{12}{c}{Geometric} \\
              \cmidrule(lr){3-14}\cmidrule(lr){15-26}
                 &
                \multicolumn{4}{r}{PSNR$_{\textrm{color}}$} & 
                \multicolumn{4}{r}{SSIM$_{\textrm{color}}$} &
                \multicolumn{4}{r}{LPIPS$_{\textrm{color}}$} &
                \multicolumn{4}{r}{MSE$_{\textrm{depth}}$} & 
                \multicolumn{4}{r}{CD$_{\textrm{mesh}}$} &
                \multicolumn{4}{r}{Comp.$_{\textrm{mesh}}@0.1m$} \\
             \cmidrule(lr){1-2}
             \cmidrule(lr){3-6}\cmidrule(lr){7-10}\cmidrule(lr){11-14}
             \cmidrule(lr){15-18}\cmidrule(lr){19-22}\cmidrule(lr){23-26}
                Cond. & Inpa.
                & 5\% & 10\% & 20\% & 50\%  
                & 5\% & 10\% & 20\% & 50\%  
                & 5\% & 10\% & 20\% & 50\%
                & 5\% & 10\% & 20\% & 50\%  
                & 5\% & 10\% & 20\% & 50\%  
                & 5\% & 10\% & 20\% & 50\%  \\
             \cmidrule(lr){1-2}
             \cmidrule(lr){3-14}
             \cmidrule(lr){15-26}
             &                  &          9.33 &          9.30 &          9.27 &          9.45  &                    0.331 &          0.330 &          0.330 &         0.333              &          0.637 &          0.636 &          0.637 &          0.635       &          1.309 &          1.304 &          1.310 &         1.293          & 3061          & 2777          & 2463          & 1758                 & 0.513          & 0.598          & 0.711          & 0.862           \\
             & $\checkmark$     &          12.4 &          14.7 &          16.5 &          17.9  &                    0.411 &          0.496 &          0.557 &         0.583              &          0.520 &          0.446 &          0.393 &          0.359       &          1.001 &          0.837 &          0.761 &         0.730          & 1934          & 850           & 443           & 149                  & 0.600          & 0.781          & 0.881          & 0.931           \\
$\checkmark$ &                  &          12.5 &          13.6 &          14.7 &          16.1  &                    0.444 &          0.473 &          0.513 &         0.556              &          0.449 &          0.409 &          0.362 & \textbf{0.315}       &          0.897 &          0.808 & \textbf{0.662} & \textbf{0.595}         & 1163          & \textbf{699}  & 176           & \textbf{99.0}        & \textbf{0.751} & 0.817          & 0.887          & 0.928           \\
$\checkmark$ & $\checkmark$     & \textbf{14.6} & \textbf{16.0} & \textbf{17.4} & \textbf{18.4}  &           \textbf{0.522} & \textbf{0.555} & \textbf{0.593} & \textbf{0.603}             & \textbf{0.448} & \textbf{0.399} & \textbf{0.359} &          0.338       & \textbf{0.825} & \textbf{0.805} &          0.688 &         0.628          & \textbf{1058} & 902           & \textbf{156}  & 100                  & 0.747          & \textbf{0.839} & \textbf{0.909} & \textbf{0.936}  \\
             \bottomrule
        \end{tabular}
    }
    \vspace{-0.3cm}
    \caption{\textbf{Ablation studies of the effects of conditioning (Cond., $\beta=1$) and inpainting (Inpa.) on visual (color images) and geometric (depth maps and meshes) results on the task of scene synthesis from sparse RGBD inputs.}
    The combined use of conditioning and inpainting leads to superior performance in visual metrics, and inpainting plays a less significant role in terms of geometric results when the model is conditioned.
    \vspace{-0.3cm}
    }
    \label{table:abla_compo}
\end{table*}
\begin{table*}[!ht]
    \centering
    \resizebox{\textwidth}{!} {
        \begin{tabular}[t]{*{25}{c}}
             \toprule
               \multirow{3}{*}{\vspace{-0.35cm}\makecell{Guidance \\ Factor $\beta$}} & \multicolumn{12}{c}{Visual} & \multicolumn{12}{c}{Geometric} \\
              \cmidrule(lr){2-13}\cmidrule(lr){14-25}
                 &
                \multicolumn{4}{c}{PSNR$_{\textrm{color}}$} & 
                \multicolumn{4}{c}{SSIM$_{\textrm{color}}$} &
                \multicolumn{4}{c}{LPIPS$_{\textrm{color}}$} &
                \multicolumn{4}{c}{MSE$_{\textrm{depth}}$} & 
                \multicolumn{4}{c}{CD$_{\textrm{mesh}}$} &
                \multicolumn{4}{c}{Comp.$_{\textrm{mesh}}@0.1m$} \\
             \cmidrule(lr){2-5}\cmidrule(lr){6-9}\cmidrule(lr){10-13}\cmidrule(lr){14-17}\cmidrule(lr){18-21}\cmidrule(lr){22-25}
                & 5\% & 10\% & 20\% & 50\%  
                & 5\% & 10\% & 20\% & 50\%  
                & 5\% & 10\% & 20\% & 50\%
                & 5\% & 10\% & 20\% & 50\%  
                & 5\% & 10\% & 20\% & 50\%  
                & 5\% & 10\% & 20\% & 50\%  \\
             \cmidrule(lr){1-1}\cmidrule(lr){2-13}\cmidrule(lr){14-25}
0.0     & 12.4          & 14.7          & 16.5          & 17.9                 & 0.411          & 0.496          & 0.557          & 0.583               & 0.520          & 0.446          & 0.393          & 0.359               &           1.001 &          0.837 &          0.761 &          0.730 &                  1934 &          850  &          443 &         149   &               0.600 &          0.781 &          0.881 &         0.931    \\
0.5     & 13.2          & 15.5          & 17.1          & 18.2                 & 0.452          & 0.530          & 0.578          & 0.596               & 0.496          & 0.418          & 0.374          & 0.347               &           0.999 &          0.845 &          0.772 &          0.719 &                  1980 &          606  &          223 &         111   &               0.653 &          0.818 &          0.894 &         0.933    \\
1.0     & \textbf{14.6} & \textbf{16.0} & 17.4          & 18.4                 & 0.522          & 0.555          & 0.593          & 0.603               & 0.448          & 0.399          & 0.359          & 0.338               &  \textbf{0.825} &          0.805 &          0.688 &          0.628 &         \textbf{1058} &          902  &          156 &         100   &               0.747 &          0.839 &          0.909 &         0.936    \\
2.0     & 14.5          & 15.8          & \textbf{17.5} & \textbf{18.4}        & \textbf{0.532} & \textbf{0.561} & \textbf{0.598} & \textbf{0.606}      & \textbf{0.439} & \textbf{0.393} & \textbf{0.352} & \textbf{0.334}      &           0.894 & \textbf{0.800} & \textbf{0.654} &          0.593 &                  1562 & \textbf{515}  & \textbf{144} &         87.2  &      \textbf{0.753} & \textbf{0.846} & \textbf{0.910} & \textbf{0.936}   \\
5.0     & 13.3          & 14.9          & 17.1          & 18.2                 & 0.488          & 0.531          & 0.579          & 0.598               & 0.475          & 0.418          & 0.367          & 0.342               &           0.992 &          0.856 &          0.663 & \textbf{0.582} &                  2551 &          1676 &          175 & \textbf{87.2} &               0.747 &          0.842 &          0.908 &         0.934    \\
             \bottomrule
        \end{tabular}
    }
    \vspace{-0.3cm}
    \caption{\textbf{Ablation studies of the effects of guidance factor $\beta$ on visual (color images) and geometric (depth maps and meshes) results on the task of scene synthesis from sparse RGBD inputs. }
    A larger $\beta$ ideally strengthens the conditioning effect. The visual metrics are found to be the best when $\beta$ is set to 1 or 2. Interestingly, the optimal value of $\beta$ for the best geometry appears to increase as views become denser. However, setting such a large value of $\beta$ (e.g. $\beta=5$) leads to sub-optimal visual results.
    \vspace{-0.3cm}
    }
    \label{table:abla_scale}
\end{table*}
\begin{table*}[h]
    \vspace{-0.3cm}
    \centering
    \resizebox{\linewidth}{!} {
        \begin{tabular}[t]{l*{24}{c}}
             \toprule
               \multirow{3}{*}{\vspace{-0.2cm}Methods} & \multicolumn{12}{c}{Visual} & \multicolumn{12}{c}{Geometric} \\
              \cmidrule(lr){2-13}\cmidrule(lr){14-25}
                 &
                \multicolumn{4}{c}{PSNR$_{\textrm{color}}$} & 
                \multicolumn{4}{c}{SSIM$_{\textrm{color}}$} &
                \multicolumn{4}{c}{LPIPS$_{\textrm{color}}$} &
                \multicolumn{4}{c}{MSE$_{\textrm{depth}}$} & 
                \multicolumn{4}{c}{CD$_{\textrm{mesh}}$} &
                \multicolumn{4}{c}{Comp.$_{\textrm{mesh}}@0.1m$} \\
             \cmidrule(lr){2-5}\cmidrule(lr){6-9}\cmidrule(lr){10-13}\cmidrule(lr){14-17}\cmidrule(lr){18-21}\cmidrule(lr){22-25}
                & 5\% & 10\% & 20\% & 50\%  
                & 5\% & 10\% & 20\% & 50\%  
                & 5\% & 10\% & 20\% & 50\%
                & 5\% & 10\% & 20\% & 50\%  
                & 5\% & 10\% & 20\% & 50\%  
                & 5\% & 10\% & 20\% & 50\%  \\
             \cmidrule(lr){1-1}\cmidrule(lr){2-13}\cmidrule(lr){14-25}
                NGP        & 10.4          & 12.4          & 14.4          & 17.4              & 0.293          & 0.377          & 0.437          & 0.498              & 0.582          & 0.476          & 0.415          & 0.376          & 7.01          & 7.39          & 7.37          & 6.52              & 29994           & 22973           & 16676          & 8066             & 0.289          & 0.446          & 0.646          & 0.850          \\
                DS-NGP     & 10.1          & 11.9          & 13.5          & 15.3              & 0.205          & 0.281          & 0.321          & 0.361              & 0.605          & 0.533          & 0.503          & 0.476          & 2.30          & 1.69          & 1.44          & 1.17              & 5362            & 1701            & 768            & 230              & 0.529          & 0.722          & 0.845          & 0.918          \\
                N-RGBD     & 14.1          & \textbf{16.5} & \textbf{18.4} & 20.0              & 0.401          & 0.513          & 0.595          & 0.652              & 0.490          & \textbf{0.384} & \textbf{0.320} & 0.287          & 1.51          & 1.23          & 1.21          & 1.13              & 3503            & 1345            & 1210           & 643              & 0.705          & 0.819          & 0.857          & 0.872          \\
                DDP        & 14.1          & 16.1          & 18.3          & \textbf{20.9}     & 0.418          & 0.504          & \textbf{0.599} & \textbf{0.702}     & 0.517          & 0.410          & 0.329          & \textbf{0.259} & 1.44          & 1.05          & 0.94          & 0.88              & 2363            & 1013            & 637            & 451              & 0.507          & 0.624          & 0.719          & 0.812          \\
                Ours       & \textbf{14.6} & 16.0          & 17.4          & 18.4              & \textbf{0.522} & \textbf{0.555} & 0.593          & 0.603              & \textbf{0.448} & 0.399          & 0.359          & 0.338          & \textbf{0.82} & \textbf{0.80} & \textbf{0.68} & \textbf{0.62}     & \textbf{1058}   & \textbf{902 }   & \textbf{156 }  & \textbf{100 }    & \textbf{0.747} & \textbf{0.839} & \textbf{0.909} & \textbf{0.936} \\
             \bottomrule
        \end{tabular}
    }
    \vspace{-0.3cm}
    \caption{\textbf{Quantitative comparison with other approaches on visual (color images) and geometric (depth maps and meshes) results on the task of scene synthesis from sparse RGBD inputs.} 
    Our approach outperforms other methods consistently in terms of geometric metrics as the sparsity varies.
    However, our method only show a definitive visual advantage across various visual metrics when the input views are highly sparse (5\%). 
    }
    \vspace{-0.3cm}
    \label{table:novel_view}
\end{table*}
%
%

%% file: tex/conclusion.tex
\vspace{-0.2cm}
\section{Discussions}
\label{sec:conclu}

\vspace{-0.2cm}
\noindent\textbf{Limitations.} Our current implementation has several limitations that may impact its usefulness in some scenarios. Firstly, it is incapable of handling color discrepancies caused by lighting variations. Secondly, it lacks surface extrapolation capabilities that can be provided through implicit field representation. Lastly, the limited receptive field of our design is confined to the observable volume of the current camera view and may result in inconsistent and discontinuous predictions, particularly in the case of a large circular camera trajectory.

\noindent\textbf{Future Works.} To improve our design, potential areas of investigation include:
(1) Modeling color smoothness and variation, as demonstrated by NeRF~\cite{NeRF}.
(2) Supporting advanced physical lighting effects, such as SVBRDF, as implemented in TANGO~\cite{TANGO}.
(3) Incorporating appearance/surface extrapolation by learning an implicit field, such as NeRF~\cite{NeRF, instantNGP} or SDF~\cite{NeuS, NeuralRGBD}.
(4) Exploring generative~\cite{GAUDI} or optimizable~\cite{NeRF--, NoPeNeRF} camera trajectories for scene synthesis.
(5) Investigating reconstruction from sparse-view RGB inputs only, using depth inpainting/estimation following the reverse sampling technique proposed in~\cite{RePaint} by utilizing a versatile RGBD diffusion model.
(6) Leveraging the multi-modal~\cite{CLIP} or generative~\cite{LDM} power of large-scale pre-trained models, such as the recently widespread Stable Diffusion~\cite{LDM}.

%% file: tex/inline-supp.tex
\begin{table*}[t]
    \centering
     {\Large \bf 
     RGBD2: Generative Scene Synthesis via Incremental \\ View Inpainting using RGBD Diffusion Models \\
     \vspace{0.3cm}
     --- Supplementary Material --- \par}
     \vspace{0.3cm}
\end{table*}


\begin{figure*}[t]
    \centering
    \includegraphics[width=\linewidth]{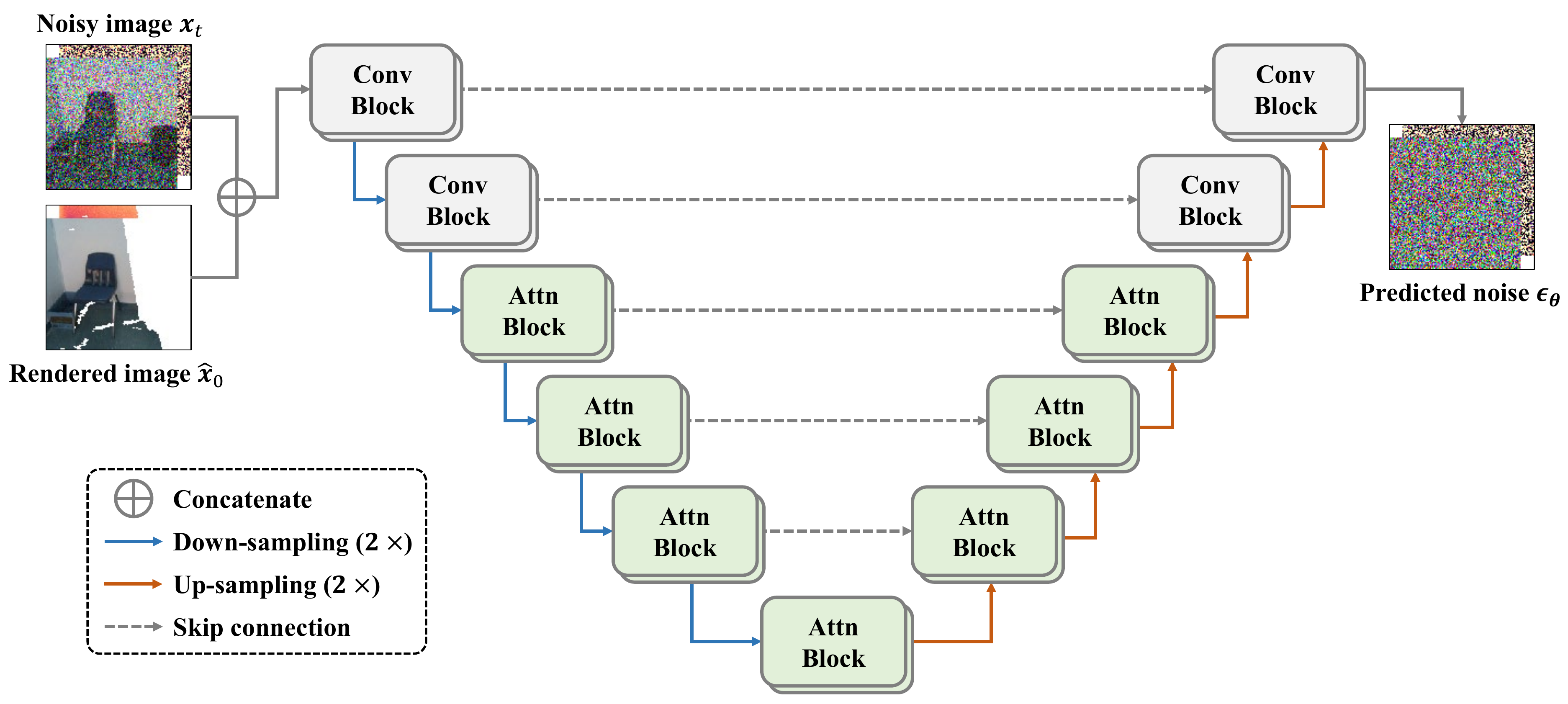}
    \caption{\textbf{The Architecture of the Utilized Diffusion Network.} 
    The network is constructed based on UNet~\cite{UNet, LDM} and takes the concatenation of the noisy image $\mathbf{x}_t$ and the rendered image $\hat{\mathbf{x}}_0$ as its input. Its objective is to predict the noise component $\epsilon_t$ added to the corrupted $\mathbf{x}_t$ by generating $\epsilon_\theta$.
    }
    \label{fig:denoisenet}
\end{figure*}

\FloatBarrier

\setcounter{section}{0} 

\vspace{-0.5cm}
\section{Other Implementation Details}
\vspace{-0.3cm}

\noindent \textbf{Network Architecture.}
Our denoising network serves the purpose of estimating the noise component $\epsilon_t$, added to the clean image $\mathbf{x}_0$, by predicting $\epsilon_\theta$. 
To construct the denoising network, we follow the conventional works\cite{DDPM, DDIM, LDM} and employ a UNet-like\cite{UNet, LDM} architecture. 
To make the most of the rich details present in $\hat{\mathbf{x}}_0$, we condition the network by concatenating it with the noisy image $\mathbf{x}_t$ to create an 8-channel input. 
The output of the network is a 4-channel prediction $\epsilon_\theta$ of the added noise $\epsilon_t$. 
For a better understanding of the network architecture, please refer to Figure~\ref{fig:denoisenet}. 
Our network operates on six different spatial resolutions, namely $128\times128$, $64\times64$, ..., and $4\times4$, by halving the last resolution via spatial down-sampling sequentially. 
To achieve this pyramid-like structure, the network is built with five down-sampling and up-sampling blocks with skip connections, which allows for the full reuse of low-level detailed information of known regions $\hat{\mathbf{x}}_0$.
In addition, each block of our network contains two residual structures, with some even incorporating attention modules to aid in the learning of long-range dependence. 
These self-attentive modules are located at spatial resolutions ranging from $32\times32$ to $4\times4$ and efficiently enhance the denoising ability by aggregating the long-range context. The dimension of each channel is set to $128$, $256$, $384$, $384$, $512$, $512$, respectively. 
Notably, we set the group number of GroupNorm in each block to one, effectively turning the group normalization to LayerNorm. 
This was done as we found a severe color-shift problem when using GroupNorm with a large number of groups ($\#\textrm{Group} \gg 1$) through experimentation.

\noindent\textbf{Comparison Details.} 
In our implementation of NGP\cite{instantNGP}, we strictly followed the official CUDA implementation, which utilized pure CUDA C++ to build the DS\cite{DS-NeRF} loss. The strength of depth supervision could be easily controlled through a weighting factor, which we set to $1.0$ for DS-NGP\cite{instantNGP, DS-NeRF} in all cases. We utilized the base configuration and trained it for $30,000$ iterations. 
For the implementation of N-RGBD\cite{NeuralRGBD} and DDP\cite{DDP}, we strictly followed their official PyTorch implementation to maintain consistency. To enable a fair comparison with other baselines, such as (DS-)NGP\cite{instantNGP, DS-NeRF}, we trained them for $30,000$ iterations as well. To determine their suitable volume range, we utilized the volume normalization parameters calculated from the ground-truth scene.

\noindent\textbf{Evaluation Details.} 
To ensure a fair comparison, we measured the color and depth metrics for rendered images. Specifically, when the method is NeRF-based\cite{NeRF}, we used the radiance field to render images and measured the metrics. In contrast, for our approach, we relied on the mesh rasterization result. 
Regarding mesh metrics, we observed that different methods employ diverse approaches (e.g. marching cubes, TSDF fusion, back-projection) to extract the mesh. To ensure fairness, we constructed all meshes through back-projection, including the ground-truth mesh. This approach helps mitigate the potential effects of unexpected factors that could impact the quantitative results. Moreover, it greatly enhances the chamfer accuracy of field-based methods such as N-RGBD\cite{NeuralRGBD} and NGP~\cite{instantNGP}. The extrapolation nature brought by implicit fields results in some extended surface (e.g. extracted by marching cubes), which could significantly raise the chamfer distance.

\noindent\textbf{Experiment Details.} 
The diffusion network operates solely in pixel space with a fixed image resolution of $128\times 128$. To maintain training efficiency, we limit the rendering chunk size to 7 images instead of constructing the scene mesh using all previously known images. This approach allows us to approximate the exact scene mesh while minimizing computational overhead.
To train with classifier-free guidance~\cite{ClassifierFree}, we randomly drop images using a dropout probability of $0.1$.
During inference, our diffusion model utilizes a DDIM~\cite{DDIM} sampler with 50 steps. In addition, we truncate edges whose lengths exceed 0.1 and eliminate faces with depth values below 0.1. To minimize complexity and reduce potential artifacts, we apply a voxel pooling operation with a voxel size of 0.02 to the resulting meshes.


\section{Limitations}
\begin{figure}[!htp]
    \centering
    \includegraphics[width=\linewidth]{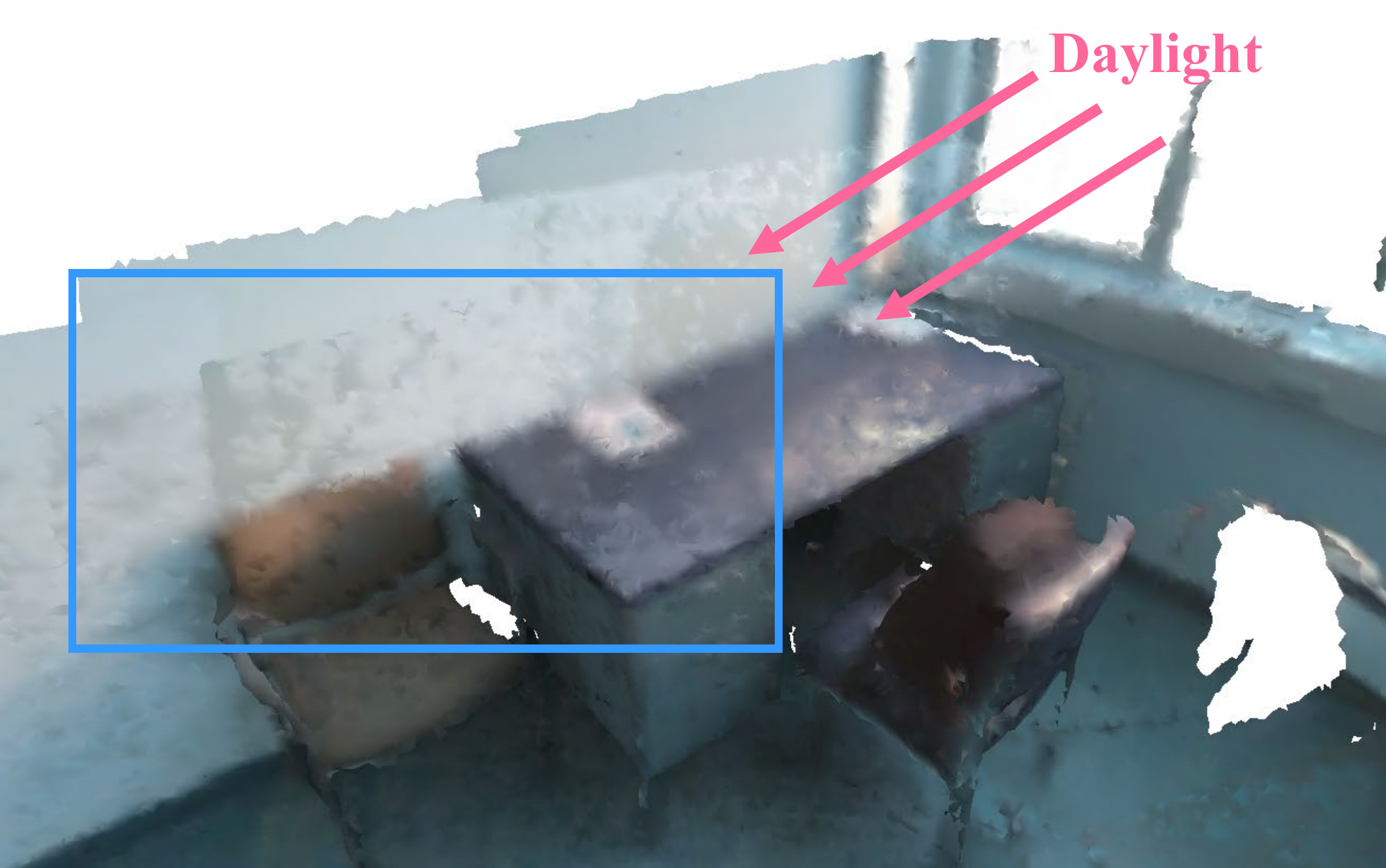}
    \caption{
    \textbf{Color Disharmony in Our Method.} It is evident that the wall and table exhibit irregular spots with uneven coloration. The discrepancy arises from outside light sources that alter the surface color depending on the camera's viewpoint. Unfortunately, our current surface modeling approach is not equipped to address this issue effectively.
    }
    \label{fig:limitation-color-disharmony}
\end{figure}

\noindent \textbf{Color Disharmony.}
Our method relies on an intermediate mesh for RGBD image rendering. To simplify the process, we combine separate meshes by concatenating all vertices and triangular faces. However, the current implementation does not consider physical lighting effects in the mesh, which could result in poor color smoothness and unpleasant visual quality, particularly when lighting variation is significant, as shown in Figure~\ref{fig:limitation-color-disharmony}. 
At present, we do not address this particular issue in our work as our main focus is on demonstrating the versatility of diffusion models for RGBD inpainting and the quality of geometric reconstruction. However, we acknowledge that the problem can be mitigated in the future through the application of color averaging techniques, assigning appropriate materials to the mesh, and modifying the shader for realistic rendering. This is an area that we intend to explore in future work.

\begin{figure}[!htp]
    \centering
    \includegraphics[width=\linewidth]{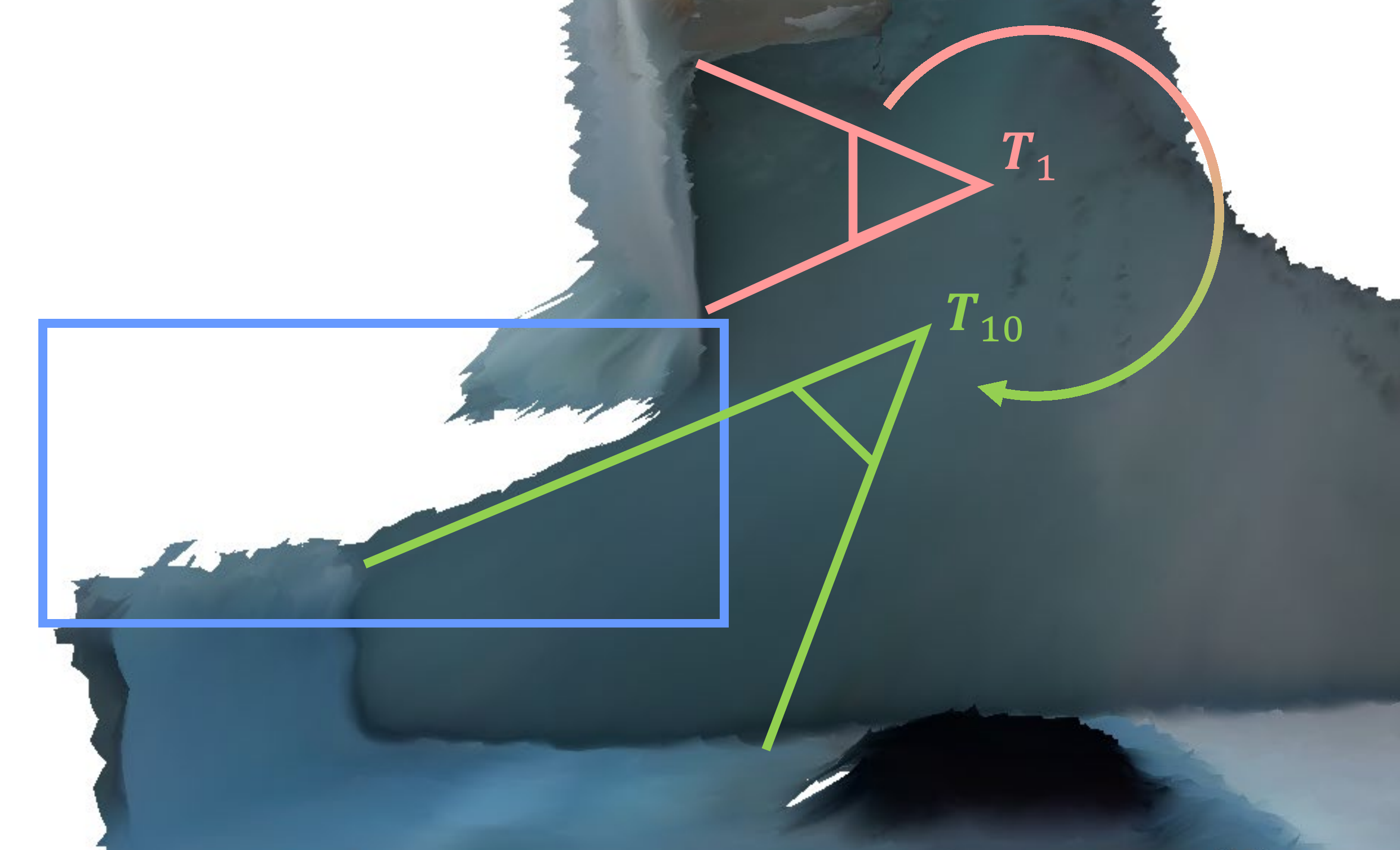}
    \caption{
    \textbf{Geometric Discontinuity Arising from Limited Receptive Field in Our Method.} When the camera moves gradually from $\mathbf{T}_1$ to $\mathbf{T}_{10}$ along a circular trajectory as depicted above, even though the current view at $\mathbf{T}_{10}$ may have already taken into account the geometry within its immediate field of view, it may not be aware of the geometry with longer-term dependencies, resulting in surface discontinuity.
    It is clear that view prediction may be subject to inconsistencies when the surface is joined from the opposite side, as opposed to being extended continuously while the camera is in motion.
    }
    \label{fig:limitation-big-loop}
\end{figure}

\noindent \textbf{Limited Receptive Volume.}
Our current design has a relatively limited receptive volume. It reconstructs missing parts by projecting the currently observed geometry into the camera viewing plane, and each view is processed independently, without knowledge of any semantic information possessed by distant parts. While this approach respects observable consistency, it cannot infer a global concept to ensure coherent predictions as the camera moves through the scene. This may result in geometric inconsistencies in some rare cases, particularly when the camera trajectory is large and circular, as shown in Figure~\ref{fig:limitation-big-loop}.
